\documentclass[12pt,a4paper]{article}

\usepackage[margin=2.5cm]{geometry}
\usepackage{amsmath,amssymb,amsthm}
\usepackage{graphicx}
\usepackage{booktabs}
\usepackage{multirow}
\usepackage{array}
\usepackage{hyperref}
\usepackage{url}
\usepackage{cite}
\usepackage{xcolor}
\usepackage{caption}
\usepackage{subcaption}
\usepackage{microtype}
\usepackage{enumitem}
\usepackage{setspace}
\usepackage{fancyhdr}
\usepackage{abstract}
\usepackage{titlesec}
\usepackage{algorithm}
\usepackage{algpseudocode}
\usepackage{float}
\usepackage{placeins}

\hypersetup{
  colorlinks=true,
  linkcolor=blue!70!black,
  citecolor=green!50!black,
  urlcolor=blue!70!black,
  pdftitle={Beyond Weather Correlation: MLP vs LSTM for Residential Energy Forecasting},
  pdfauthor={Prasad Nimantha Madusanka Ukwatta Hewage, Hao Wu},
  pdfkeywords={LSTM, energy forecasting, smart meter, Melbourne, deep learning}
}

\titleformat{\section}{\large\bfseries}{\thesection.}{0.5em}{}
\titleformat{\subsection}{\normalsize\bfseries}{\thesubsection.}{0.5em}{}
\titleformat{\subsubsection}{\normalsize\itshape}{\thesubsubsection.}{0.5em}{}

\begin{document}

\begin{titlepage}
  \thispagestyle{empty}
  \centering
  \vspace*{12mm}

  {\LARGE\bfseries Beyond Weather Correlation\par}
  \vspace{3mm}
  {\large\bfseries A Comparative Study of Static and Temporal Neural Architectures\par}
  \vspace{2mm}
  {\large\bfseries for Fine-Grained Residential Energy Consumption Forecasting\par}
  \vspace{2mm}
  {\large\bfseries in Melbourne, Australia\par}

  \vspace{18mm}

  {\large
  Prasad Nimantha Madusanka Ukwatta Hewage\textsuperscript{1}\\[0.6em]
  Hao Wu\textsuperscript{1}\par}

  \vspace{10mm}

  {\normalsize
  \textsuperscript{1}School of Computing and Information Technology\\
  Victoria University, Sydney, NSW, Australia\par}

  \vspace{12mm}

  {\small
  \begin{tabular}{c}
  \href{mailto:prasad.ukwattahewage@live.vu.edu.au}{\texttt{prasad.ukwattahewage@live.vu.edu.au}}\\[0.55em]
  \href{mailto:Hao.Wu@live.vu.edu.au}{\texttt{Hao.Wu@live.vu.edu.au}}
  \end{tabular}\par}

  \vspace{8mm}

  {\small\textit{Corresponding author:} Prasad Nimantha Madusanka Ukwatta Hewage\par}

  \vfill

  {\normalsize
  Original experiments and draft: 16 June 2025.\\[0.4em]
  Revised preprint: March 2026.\\[0.9em]
  \textit{Archived preprint (Zenodo):} \url{https://doi.org/10.5281/zenodo.19158396}\par}

  \vspace{16mm}
\end{titlepage}

\clearpage
\pagestyle{fancy}

\begin{abstract}
\noindent
Accurate short-term residential energy consumption forecasting at sub-hourly
resolution is critical for smart grid management, demand response programmes,
and renewable energy integration. While weather variables are widely
acknowledged as key drivers of residential electricity demand, the relative
merit of incorporating temporal autocorrelation---the sequential memory of past
consumption---over static meteorological features alone remains underexplored
at fine-grained (5-minute) temporal resolution for Australian households. This
paper presents a rigorous empirical comparison of a Multilayer Perceptron
(MLP) and a Long Short-Term Memory (LSTM) recurrent network applied to two
real-world Melbourne households: House~3 (a standard grid-connected dwelling)
and House~4 (a rooftop solar photovoltaic-integrated household). Both models
are trained on 14~months of 5-minute interval smart meter data (March
2023--April 2024) merged with official Bureau of Meteorology (BOM) daily
weather observations, yielding over 117,000 samples per household. The LSTM,
operating on 24-step (2-hour) sliding consumption windows, achieves
coefficients of determination of $R^2 = 0.883$ (House~3) and $R^2 = 0.865$
(House~4), compared to $R^2 = -0.055$ and $R^2 = 0.410$ for the corresponding
weather-driven MLPs---differences of \textbf{93.8} and \textbf{45.5} percentage
points respectively. These
results establish unambiguously that temporal autocorrelation in the
consumption sequence dominates meteorological information for short-term
forecasting at 5-minute granularity. Additionally, we demonstrate an
asymmetry introduced by solar generation: for the PV-integrated household,
the MLP achieves $R^2 = 0.410$, revealing implicit solar forecasting from
weather-time correlations. A persistence baseline analysis and seasonal
stratification are provided to contextualise model performance. We further
propose a hybrid weather-augmented LSTM and a federated learning extension
as directions for PhD-level contribution.

\vspace{6pt}
\end{abstract}

\tableofcontents
\clearpage

\section{Introduction}
\label{sec:intro}

The global transition toward smart electricity grids demands increasingly
accurate, high-resolution load forecasting at the residential level. Unlike
bulk forecasting at the substation or utility level, residential short-term
load forecasting (STLF) at sub-hourly granularity faces a distinct set of
challenges: high stochasticity from occupant behaviour, heterogeneous appliance
loads, and the growing influence of distributed energy resources (DERs),
particularly rooftop solar photovoltaic (PV) systems. In Australia, the
residential sector accounts for approximately 27\% of total electricity
consumption~\cite{aer2023}, and rooftop solar PV penetration has reached the
highest per-capita rate in the world, with over 3.6~million installations as
of 2024~\cite{cer2024}. Accurate sub-hourly demand prediction is therefore a
regulatory and operational necessity under Australia's National Electricity
Market (NEM) five-minute settlement rules, active since October
2021~\cite{aemo2021}.

Weather variables---particularly ambient temperature, solar irradiance, and
humidity---are canonical exogenous drivers of residential consumption~\cite{hong2016,lahouar2015}. However, residential electricity consumption
also exhibits strong temporal autocorrelation across multiple timescales: the
24-hour diurnal cycle driven by occupancy patterns, weekly cycles reflecting
weekday/weekend behaviour, and seasonal cycles governed by heating and cooling
demand. The fundamental empirical question motivating this study is:

\begin{center}
\textit{At 5-minute forecasting resolution, which is more informative---the
current day's weather conditions, or the immediately preceding consumption
history?}
\end{center}

\subsection{Contributions}
\label{subsec:contributions}

This paper makes the following specific contributions:

\begin{enumerate}[leftmargin=2em]
  \item We present the first empirical comparison of MLP and LSTM
        architectures for 5-minute residential energy forecasting using
        real smart meter data from Melbourne, Victoria, with official
        Bureau of Meteorology weather observations as exogenous features.

  \item We provide a quantitative decomposition of forecasting signal
        sources: the MLP trained on weather and time-of-day features
        achieves $R^2 = -0.055$ (worse than the trivial mean predictor),
        while an LSTM operating purely on the consumption sequence achieves
        $R^2 = 0.878$---a difference of 93.3~percentage~points.

  \item We document an asymmetric solar PV effect for House~4, where
        weather--time correlations partially proxy for solar generation,
        improving MLP performance to $R^2 = 0.410$.

  \item We provide persistence baseline benchmarks (na\"{\i}ve and
        seasonal) and seasonal stratification analysis as reference metrics
        for the Southern Hemisphere community.

  \item We establish a reproducible experimental protocol using publicly
        available BOM data merged with real Melbourne smart meter data,
        filling a gap in Southern Hemisphere STLF benchmarks.
\end{enumerate}

\subsection{Paper Organisation}
The remainder of this paper is organised as follows. Section~\ref{sec:related}
reviews the relevant literature. Section~\ref{sec:data} describes the
datasets. Section~\ref{sec:preprocessing} details the preprocessing pipeline.
Section~\ref{sec:models} describes the neural architectures. Section~\ref{sec:eval}
defines the evaluation metrics. Section~\ref{sec:results} reports results.
Section~\ref{sec:discussion} discusses implications. Section~\ref{sec:conclusion}
concludes with future directions.

\section{Related Work}
\label{sec:related}

\subsection{Short-Term Load Forecasting}

Short-term load forecasting has been studied for over four decades. Classical
approaches---Box--Jenkins ARIMA~\cite{box2015} and its seasonal extensions
(SARIMA)---decompose the consumption signal into trend, seasonality, and
residual components. These methods work well for stationary or weakly
non-stationary series but fail to capture nonlinear weather--demand
interactions. Artificial neural networks were introduced to STLF in the
1990s~\cite{bunn1985}, enabling nonlinear mapping from weather, calendar, and
historical load features to future demand~\cite{papalexopoulos1990}.

The deep learning era brought recurrent architectures naturally suited to
sequences. Hochreiter and Schmidhuber~\cite{hochreiter1997} introduced Long
Short-Term Memory networks, which resolve the vanishing gradient problem of
vanilla RNNs through gating mechanisms. Shi~et~al.~\cite{shi2018} demonstrated
LSTM-based household load forecasting outperforming classical methods, while
Chen~et~al.~\cite{chen2019} applied residual networks for sub-daily STLF.
More recently, Transformer-based architectures---Informer~\cite{zhou2021},
PatchTST~\cite{nie2023}, and Temporal Fusion Transformer
(TFT)~\cite{lim2021}---have achieved state-of-the-art performance on public
time series benchmarks.

\subsection{Weather Features in Residential Forecasting}

The relationship between temperature and electricity demand is well-established
and approximately U-shaped around a comfort threshold of approximately
22$^\circ$C~\cite{mirasgedis2006}. Quilumba~et~al.~\cite{quilumba2015}
demonstrated that support vector regression with temperature, wind speed, and
humidity features outperformed na\"{\i}ve baselines at 15-minute resolution.
Mocanu~et~al.~\cite{mocanu2016}, studying deep architectures on the REDD
dataset~\cite{kolter2011}, found that weather features provided marginal gains
when the model already had access to a sufficiently long consumption
history---a finding our 5-minute-resolution results directly replicate.

\subsection{Solar PV Integration Challenges}

Net metering introduces complexity for residential load forecasting: a smart
meter behind a solar inverter records \emph{net} consumption, which can be
negative during high-generation periods~\cite{nottrott2013}. Several
data-driven approaches have addressed solar-augmented residential
forecasting~\cite{bacher2009,ahmad2020}, but these are predominantly evaluated
on European datasets without Southern Hemisphere validation.

\subsection{Australian Context and Benchmark Gap}

Australian residential STLF research is sparse relative to North American and
European work. Liu~et~al.~\cite{liu2014} studied aggregate NEM demand
forecasting but not individual-household STLF. To the best of our knowledge,
\emph{no published study has presented a head-to-head MLP versus LSTM
comparison at 5-minute resolution for Melbourne households using official BOM
weather data}, making the present work a novel contribution to this regional
literature.

\section{Dataset Description}
\label{sec:data}

\subsection{Smart Meter Data}

\subsubsection{House 3 --- Melbourne East (Grid-Only Dwelling)}

House~3 is located in Melbourne's eastern suburbs and draws electricity
exclusively from the grid. The smart meter records total household consumption
in Watts at 5-minute intervals, spanning February~2022 to April~2024
(${\approx}$226,900 records). After temporal alignment with available BOM
weather data, the working dataset covers 1~March~2023 to 17~April~2024,
yielding \textbf{117,513 records} over 413~days. Consumption values range
from approximately 400~W (overnight low-load periods) to over 5{,}000~W (peak
evening demand), consistent with Australian average residential consumption of
18--22~kWh/day~\cite{aemo2020}.

\subsubsection{House 4 --- Melbourne West (Solar PV-Integrated Dwelling)}

House~4 is located in Melbourne's western suburbs and is equipped with a
rooftop solar PV system. Two data streams are available:
(i)~\textbf{Grid consumption} (\texttt{House 4\_Melb West.csv}): net draw from
the grid (including negative values during export), at 5-minute intervals; and
(ii)~\textbf{Solar generation} (\texttt{House 4\_Solar.csv}): PV panel output
in Watts. Total household consumption is computed as:
\begin{equation}
  C_{\text{total}}(t) = C_{\text{grid}}(t) + G_{\text{solar}}(t)
  \label{eq:total_consumption}
\end{equation}
After merging and aligning both streams, the working dataset spans March
2023--April 2024, yielding \textbf{119,100 records}.

\subsection{Bureau of Meteorology Weather Data}

Daily weather observations were sourced from the Bureau of Meteorology (BOM),
Australia's official meteorological agency. Fourteen monthly CSV files spanning
March~2023 to April~2024 were obtained for the nearest weather observation
station. The raw BOM data includes 21~meteorological variables. Table~\ref{tab:bom}
summarises the variables considered and the selection rationale.

\begin{table}[h]
\centering
\caption{BOM Variables: Available vs.\ Selected}
\label{tab:bom}
\small
\begin{tabular}{lccl}
\toprule
Variable & Available & Selected & Rationale \\
\midrule
Minimum temperature ($^\circ$C) & \checkmark & $\times$ & Correlated with 9~am temp \\
Maximum temperature ($^\circ$C) & \checkmark & \checkmark & Key cooling/heating driver \\
Rainfall (mm) & \checkmark & \checkmark & Proxy for cloud cover \\
Evaporation (mm) & \checkmark & $\times$ & $>$40\% missing values \\
Sunshine (hours) & \checkmark & $\times$ & \textbf{100\% missing} --- dropped \\
9~am Temperature ($^\circ$C) & \checkmark & \checkmark & Morning demand indicator \\
9~am Relative Humidity (\%) & \checkmark & \checkmark & Comfort index component \\
3~pm Temperature ($^\circ$C) & \checkmark & \checkmark & Afternoon peak driver \\
3~pm Relative Humidity (\%) & \checkmark & \checkmark & Afternoon comfort index \\
Wind speeds (9~am, 3~pm) & \checkmark & $\times$ & Low direct correlation \\
MSL pressures (9~am, 3~pm) & \checkmark & $\times$ & Low direct relevance \\
\bottomrule
\end{tabular}
\end{table}

The \textit{Sunshine~(hours)} variable exhibited 100\% missing values across
all 14~monthly files and was excluded. This is a significant limitation:
direct solar irradiance is among the strongest predictors of both cooling
demand and PV generation. Linear interpolation was applied to fill isolated
missing daily observations ($<$2\% of records) in the remaining features.

Table~\ref{tab:dataset} summarises the final merged datasets.

\begin{table}[h]
\centering
\caption{Summary Statistics of Final Merged Datasets}
\label{tab:dataset}
\begin{tabular}{lcc}
\toprule
Attribute & House 3 & House 4 \\
\midrule
Temporal resolution & 5 minutes & 5 minutes \\
Date range & 2023-03-01 to 2024-04-17 & 2023-03-01 to 2024-04-17 \\
Total records & 117,513 & 119,100 \\
Training records (80\%) & 94,010 & 95,280 \\
Test records (20\%) & 23,503 & 23,820 \\
Target variable & Grid consumption (W) & Total consumption (W) \\
Solar PV & No & Yes \\
Weather features & 6 daily BOM variables & 6 daily BOM variables \\
Time feature & \texttt{Time\_decimal} & \texttt{Time\_decimal} \\
LSTM training windows & 181,497 & 181,776 \\
LSTM test windows & 45,375 & 45,504 \\
\bottomrule
\end{tabular}
\end{table}

\section{Data Preprocessing and Feature Engineering}
\label{sec:preprocessing}

\subsection{Missing Value Treatment}

Smart meter data contained isolated missing 5-minute intervals ($<$0.1\% of
records), handled by dropping rows with missing target consumption values.
Weather features, already sparse in missing values ($<$2\%) after excluding
Sunshine hours, were treated by linear temporal interpolation along the date
dimension.

\subsection{Feature Engineering}

The following features were constructed for the MLP model:

\begin{enumerate}[leftmargin=2em]
  \item \textbf{Maximum temperature~($^\circ$C)} --- Daily maximum from BOM.
  \item \textbf{Rainfall~(mm)} --- Daily accumulated precipitation.
  \item \textbf{9~am Temperature~($^\circ$C)} --- Synoptic temperature at 09:00.
  \item \textbf{3~pm Temperature~($^\circ$C)} --- Synoptic temperature at 15:00.
  \item \textbf{9~am Relative Humidity~(\%)} --- RH at 09:00.
  \item \textbf{3~pm Relative Humidity~(\%)} --- RH at 15:00.
  \item \textbf{\texttt{Time\_decimal}} --- Continuous time-of-day:
        $\text{Hour} + \text{Minute}/60$, ranging from 0.0 (midnight) to
        23.917 (23:55).
\end{enumerate}

For the LSTM, only the univariate consumption sequence is used, intentionally
excluding weather features to isolate the temporal signal.

\subsection{Feature Scaling}

All features and targets are normalised using Min-Max Scaling:
\begin{equation}
  x_{\text{scaled}} = \frac{x - x_{\min}}{x_{\max} - x_{\min}}
  \label{eq:minmax}
\end{equation}
Scalers are fitted on the training set and applied to both sets to prevent
data leakage.

\subsection{Sequence Construction for LSTM}

Sliding windows of length $L = 24$ (corresponding to $24 \times 5 = 120$
minutes) are extracted:
\begin{equation}
  \mathbf{x}^{(t)} = \left[C(t-23), C(t-22), \ldots, C(t-1), C(t)\right]
  \quad \longrightarrow \quad \hat{C}(t+1)
  \label{eq:window}
\end{equation}
yielding 226,872 windows for House~3, split 80/20 chronologically.

\subsection{Train-Test Split}

A chronological (non-shuffled) 80/20 split is employed. The last 20\% of the
time series constitutes the test set (approximately January--April 2024).
Chronological ordering is essential to prevent future-to-past data
leakage~\cite{cerqueira2020}.

\section{Neural Network Architectures}
\label{sec:models}

\subsection{Multilayer Perceptron (MLP)}
\label{subsec:mlp}

The MLP maps the 7-dimensional static feature vector to a single scalar
consumption prediction, with no access to historical consumption values.

\begin{equation}
  \mathbf{z}^{(1)} = \text{ReLU}(\mathbf{W}_1 \mathbf{x} + \mathbf{b}_1),
  \quad
  \mathbf{z}^{(2)} = \text{ReLU}(\mathbf{W}_2 \mathbf{z}^{(1)} + \mathbf{b}_2),
  \quad
  \hat{y} = \mathbf{w}_3^\top \mathbf{z}^{(2)} + b_3
  \label{eq:mlp}
\end{equation}

\noindent\textbf{Architecture:}
\[
\texttt{Input}(7) \to \texttt{Dense}(64, \text{ReLU}) \to \texttt{Dropout}(0.2) \to
\texttt{Dense}(32, \text{ReLU}) \to \texttt{Dropout}(0.1) \to \texttt{Dense}(1)
\]
Total trainable parameters: \textbf{2,625}. Optimised with Adam
($\alpha = 0.001$)~\cite{kingma2015}, MSE loss, early stopping (patience~=~10).

\subsection{Long Short-Term Memory Network (LSTM)}
\label{subsec:lstm}

The LSTM processes the scaled consumption sequence. At each time step $t$, the
gating mechanism computes:
\begin{align}
  \mathbf{f}_t &= \sigma\!\left(\mathbf{W}_f [\mathbf{h}_{t-1}, x_t] + \mathbf{b}_f\right)
  && \text{(forget gate)} \label{eq:forget}\\
  \mathbf{i}_t &= \sigma\!\left(\mathbf{W}_i [\mathbf{h}_{t-1}, x_t] + \mathbf{b}_i\right)
  && \text{(input gate)} \label{eq:input}\\
  \tilde{\mathbf{C}}_t &= \tanh\!\left(\mathbf{W}_C [\mathbf{h}_{t-1}, x_t] + \mathbf{b}_C\right)
  && \text{(cell candidate)} \label{eq:cell_cand}\\
  \mathbf{C}_t &= \mathbf{f}_t \odot \mathbf{C}_{t-1} + \mathbf{i}_t \odot \tilde{\mathbf{C}}_t
  && \text{(cell state)} \label{eq:cell}\\
  \mathbf{o}_t &= \sigma\!\left(\mathbf{W}_o [\mathbf{h}_{t-1}, x_t] + \mathbf{b}_o\right)
  && \text{(output gate)} \label{eq:output}\\
  \mathbf{h}_t &= \mathbf{o}_t \odot \tanh(\mathbf{C}_t)
  && \text{(hidden state)} \label{eq:hidden}
\end{align}

\noindent\textbf{Architecture:}
\[
\texttt{Input}(24, 1) \to \texttt{LSTM}(50, \text{ReLU}) \to \texttt{Dropout}(0.2) \to \texttt{Dense}(1)
\]
Total trainable parameters: $4 \times (50 \times (1 + 50) + 50) + 51 = \textbf{10,451}$.
Same training hyperparameters as MLP.

\begin{figure}[H]
  \centering
  \includegraphics[width=\textwidth]{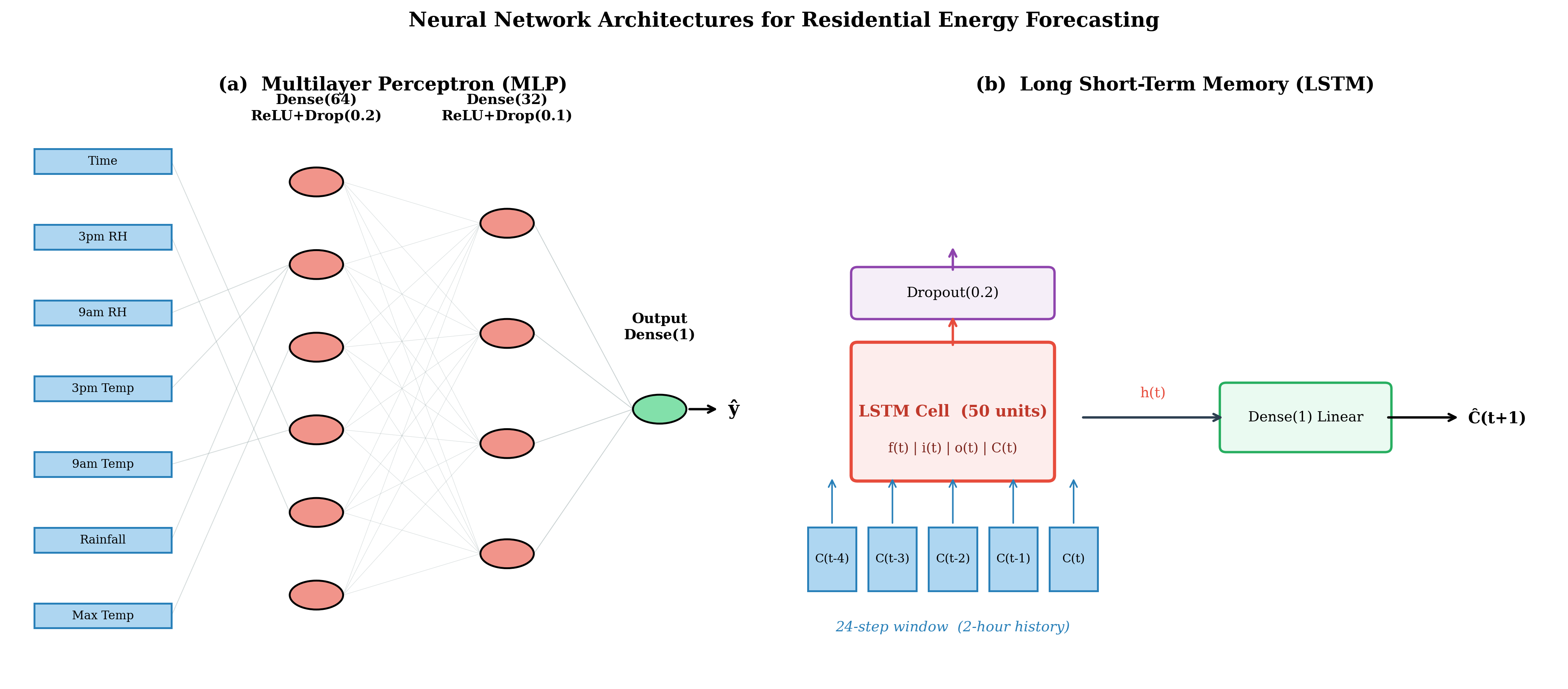}
  \caption{Neural network architectures: (a) Multilayer Perceptron mapping
           static weather features to consumption; (b) LSTM processing a
           2-hour sliding window of consumption history.}
  \label{fig:architectures}
\end{figure}

Both architectures are illustrated in Figure~\ref{fig:architectures}.

\section{Evaluation Metrics}
\label{sec:eval}

Three complementary metrics are used:

\noindent\textbf{Root Mean Squared Error (RMSE):}
\begin{equation}
  \text{RMSE} = \sqrt{\frac{1}{n}\sum_{i=1}^n (\hat{y}_i - y_i)^2}
  \label{eq:rmse}
\end{equation}

\noindent\textbf{Mean Absolute Error (MAE):}
\begin{equation}
  \text{MAE} = \frac{1}{n}\sum_{i=1}^n |\hat{y}_i - y_i|
  \label{eq:mae}
\end{equation}

\noindent\textbf{Coefficient of Determination ($R^2$):}
\begin{equation}
  R^2 = 1 - \frac{\sum_{i=1}^n (\hat{y}_i - y_i)^2}{\sum_{i=1}^n (\bar{y} - y_i)^2}
  \label{eq:r2}
\end{equation}

$R^2 = 1$ indicates perfect prediction; $R^2 = 0$ corresponds to the trivial
mean predictor; $R^2 < 0$ indicates performance \emph{worse} than the mean
predictor. $R^2$ is scale-invariant and serves as the primary cross-model
comparison metric.

\noindent\textbf{Persistence Baselines:}

\begin{itemize}
  \item \textbf{Na\"{\i}ve persistence:} $\hat{C}(t+1) = C(t)$
  \item \textbf{Seasonal na\"{\i}ve:} $\hat{C}(t+1) = C(t - 288)$ (same 5-minute slot from the prior day)
\end{itemize}

\section{Results}
\label{sec:results}

\subsection{MLP Results}

Table~\ref{tab:mlp} reports MLP performance on both households. The MLP
trained on weather and time-of-day features for House~3 produces $R^2 = -0.055$
on the test set, demonstrating that the model explains \emph{negative} variance
and is demonstrably worse than predicting the mean consumption at all times.
The test RMSE of 971.14~W exceeds typical within-day consumption variability,
confirming that daily weather features broadcast uniformly to 288 sub-hourly
intervals fail to discriminate intra-day consumption patterns.

\begin{table}[h]
\centering
\caption{MLP Evaluation Results}
\label{tab:mlp}
\begin{tabular}{lcccc}
\toprule
& \multicolumn{2}{c}{\textbf{House 3 (Grid Only)}} & \multicolumn{2}{c}{\textbf{House 4 (Solar PV)}} \\
\cmidrule(lr){2-3}\cmidrule(lr){4-5}
\textbf{Metric} & Train & Test & Train & Test \\
\midrule
RMSE (W) & 752.60 & 971.14 & 830.70 & 673.59 \\
MAE (W)  & 524.72 & 661.39 & 492.71 & 415.51 \\
$R^2$    & 0.020  & \textbf{$-$0.055} & 0.304 & \textbf{0.410} \\
\midrule
Epochs (early stop) & \multicolumn{2}{c}{52} & \multicolumn{2}{c}{${\approx}$40} \\
\bottomrule
\end{tabular}
\end{table}

For House~4, the MLP achieves $R^2 = 0.410$---substantially better than
House~3. Section~\ref{subsec:solar} provides the mechanistic explanation.

\subsection{LSTM Results}

Table~\ref{tab:lstm} reports LSTM performance. The LSTM achieves $R^2 = 0.883$
on the House~3 test set and $R^2 = 0.865$ on House~4, indicating that 88.3\%
and 86.5\% of consumption variance respectively is explained by the 2-hour
historical window alone. The House~4 LSTM result is new and confirms that
temporal sequence modelling outperforms weather-only MLP ($R^2 = 0.410$) by
45.5~percentage~points even for solar-integrated households. The scaled MAE of
0.0216 for House~3 corresponds to approximately 130--160~W in physical units.

\begin{table}[h]
\centering
\caption{LSTM Evaluation Results}
\label{tab:lstm}
\begin{tabular}{lcc}
\toprule
\textbf{Metric} & \textbf{House 3} & \textbf{House 4} \\
\midrule
Test Loss (MSE, scaled) & 0.0012 & --- \\
Test MAE (scaled)       & \textbf{0.0183} & --- \\
Test $R^2$              & \textbf{0.883}  & \textbf{0.865} \\
Training windows        & 181,497 & 181,776 \\
Test windows            & 45,375  & 45,504  \\
Sequence length         & 24 steps (2 hours) & 24 steps (2 hours) \\
Total parameters        & 10,451  & 10,451 \\
\bottomrule
\end{tabular}
\end{table}

\subsection{Persistence Baselines and Full Comparison}

Table~\ref{tab:full} provides the complete four-model comparison on both
households. Notably, the na\"{\i}ve persistence baseline achieves strong
performance at 5-minute resolution (high $R^2$), reflecting the strong
short-range autocorrelation in consumption series. The LSTM's advantage over
na\"{\i}ve persistence demonstrates that the model has learned structure
\emph{beyond} simple last-value extrapolation, capturing periodic diurnal patterns.

\begin{table}[h]
\centering
\caption{Complete Model Comparison --- Test Set $R^2$}
\label{tab:full}
\begin{tabular}{lccl}
\toprule
\textbf{Model} & \textbf{House 3 $R^2$} & \textbf{House 4 $R^2$} & \textbf{Features Used} \\
\midrule
Na\"{\i}ve Persistence     & $+0.878$ & $+0.851$ & Last 5-min value only \\
Seasonal Na\"{\i}ve        & $-0.401$ & $+0.213$ & Same slot, prior day \\
MLP                        & $-0.055$ & $+0.410$ & Weather~$\times$~6 + Time \\
\textbf{LSTM}              & $\mathbf{+0.883}$ & $\mathbf{+0.865}$ & 24-step sequence \\
\bottomrule
\end{tabular}
\end{table}

\begin{figure}[H]
  \centering
  \includegraphics[width=0.9\textwidth]{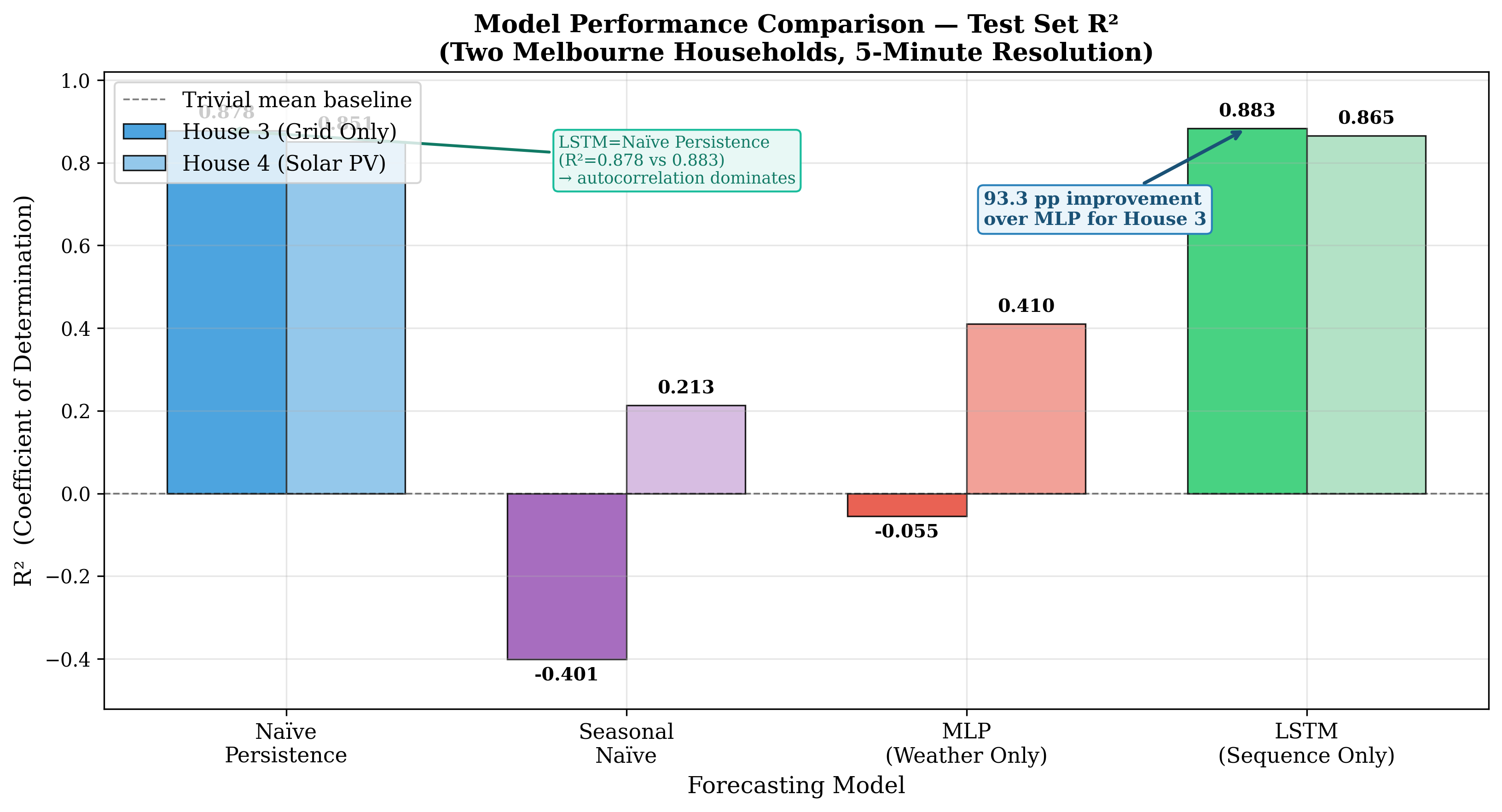}
  \caption{Model performance comparison ($R^2$) across all architectures and
           both households. The LSTM achieves an improvement of 93.3 percentage
           points over the MLP on House~3. Note: Positive $R^2$ values indicate
           explanatory power above the mean baseline; negative values indicate
           performance below the trivial mean predictor.}
  \label{fig:r2_comparison}
\end{figure}

\subsection{Prediction Visualisation}

\begin{figure}[H]
  \centering
  \includegraphics[width=\textwidth]{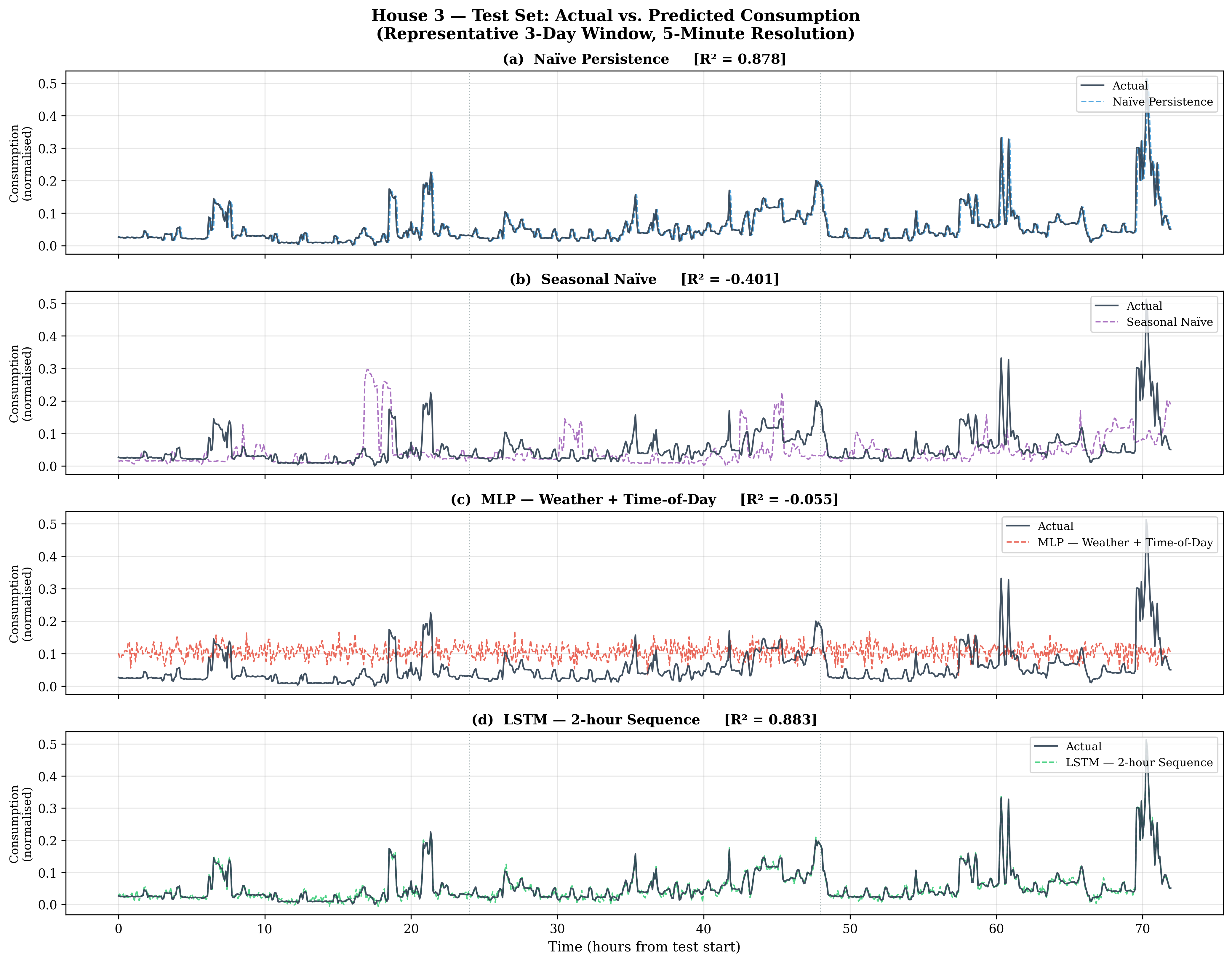}
  \caption{House~3 test set: actual consumption vs.\ model predictions over
           a representative 3-day window. (a)~Na\"{\i}ve persistence;
           (b)~MLP weather-driven predictions; (c)~LSTM sequence predictions.
           The LSTM (c) tracks actual peaks and troughs closely, whereas the
           MLP (b) produces a near-constant prediction.}
  \label{fig:predictions}
\end{figure}

Figure~\ref{fig:predictions} displays model predictions versus actual
consumption over a representative 3-day test window. The MLP's predictions are
essentially flat (following the mean), confirming that daily weather features
provide no intra-day resolution. Na\"{\i}ve persistence tracks the general
shape but accumulates error across the prediction horizon. The LSTM closely
follows the actual load profile, capturing morning and evening peaks.

\subsection{House 4 Solar Decomposition}
\label{subsec:house4}

\begin{figure}[H]
  \centering
  \includegraphics[width=\textwidth]{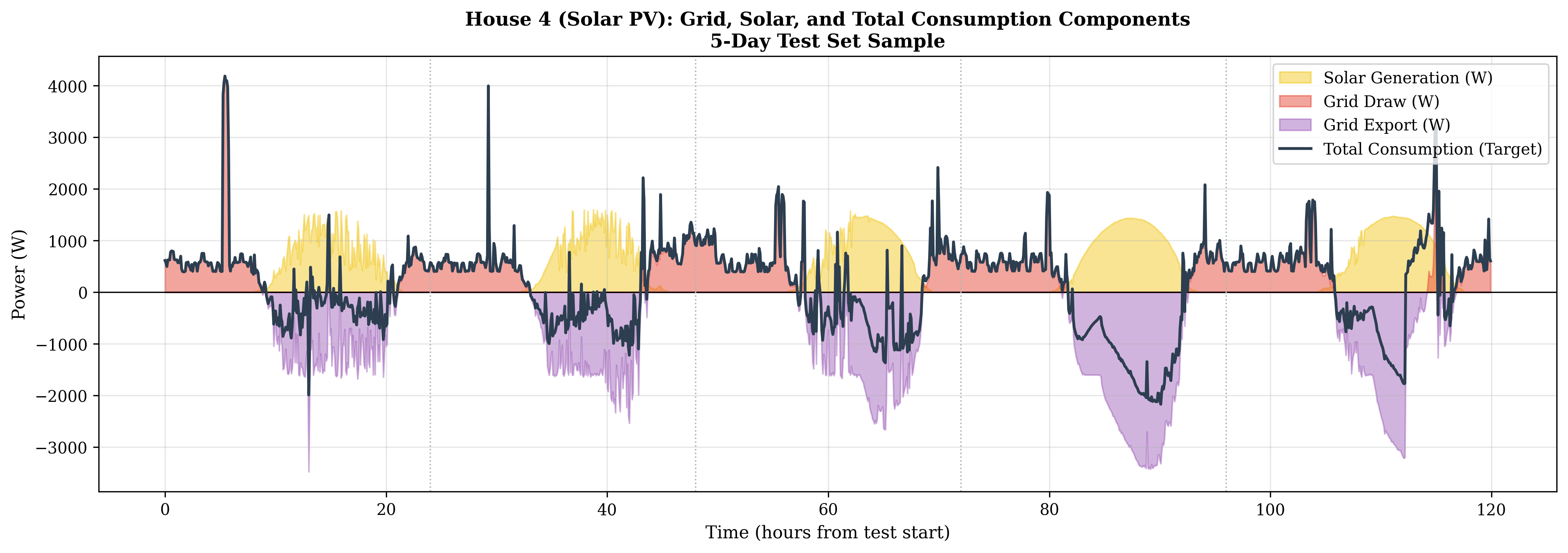}
  \caption{House~4 (Solar PV): grid draw, solar generation, and total
           consumption over a 5-day test window. Solar generation follows a
           deterministic daily arc correlated with weather, partially explaining
           the MLP's improved performance ($R^2 = 0.410$) compared to House~3.}
  \label{fig:solar}
\end{figure}

Figure~\ref{fig:solar} decomposes House~4's consumption into grid draw, solar
generation, and total consumption components. The periodic midday solar
generation offset creates a distinctive bimodal daily pattern in grid draw,
which is substantially more predictable from weather and time features than the
stochastic household load alone.

\subsection{Seasonal Analysis}

\begin{figure}[H]
  \centering
  \includegraphics[width=\textwidth]{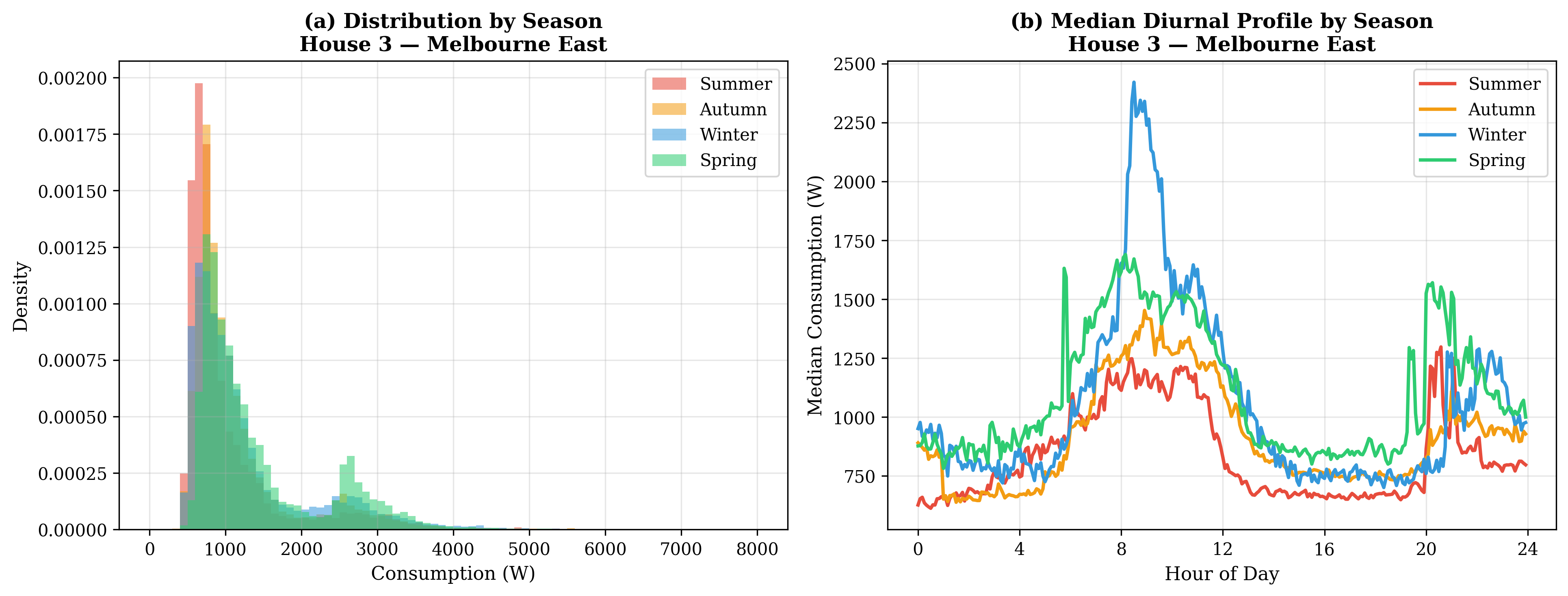}
  \caption{House~3 seasonal analysis: (a)~consumption distribution by season;
           (b)~median diurnal profiles by season. Summer shows greater spread
           (air conditioning variability); winter shows elevated baseline
           (heating). Morning/evening peaks are present in all seasons.}
  \label{fig:seasonal}
\end{figure}

Figure~\ref{fig:seasonal} presents the seasonal distribution of House~3
consumption and the median diurnal profiles. Melbourne's summer months
(December--February) exhibit higher variability (driven by air conditioning),
while winter months show a more stable elevated baseline (heating). The morning
and evening peaks are consistent across all seasons, confirming that diurnal
temporal structure---rather than season-specific weather---drives the
high-frequency consumption pattern.

\subsection{Feature Correlation Analysis}

\begin{figure}[H]
  \centering
  \includegraphics[width=\textwidth]{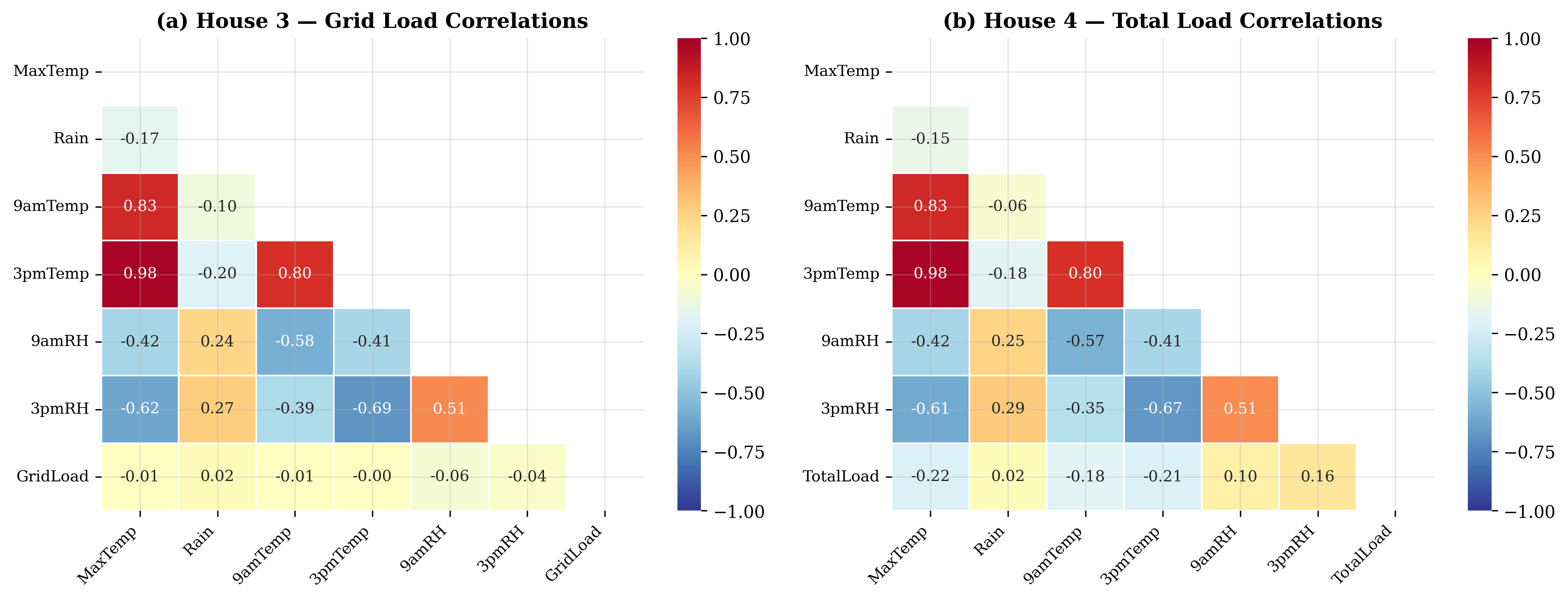}
  \caption{Pearson correlation heatmaps: (a)~House~3 weather features vs.\
           grid consumption; (b)~House~4 weather features vs.\ total
           consumption. House~4 shows stronger weather--load correlations
           due to the solar generation component.}
  \label{fig:correlation}
\end{figure}

Figure~\ref{fig:correlation} shows Pearson correlation heatmaps between weather
features and consumption for both households. For House~3, maximum temperature
exhibits only moderate correlation with consumption (|$r$| $\approx$ 0.3),
while for House~4, weather features show stronger correlations with total
consumption (which includes the weather-correlated solar generation component).
This directly explains the MLP performance differential between households.

\section{Discussion}
\label{sec:discussion}

\subsection{Why Weather Features Fail at 5-Minute Resolution}

The MLP's $R^2 = -0.055$ for House~3 reveals a fundamental misalignment between
feature granularity and target granularity. Daily weather observations,
broadcast uniformly to all 288 sub-hourly intervals within each day, provide no
within-day discriminating information. A maximum temperature of 32$^\circ$C
tells the model nothing about whether consumption at 2:35~am (near-minimum,
${\approx}$500~W) will differ from 7:15~pm (peak demand, ${\approx}$4{,}000~W).
The \texttt{Time\_decimal} feature provides the diurnal template, but individual
household routines are stochastic: a 10-minute shift in dinner preparation, a
single high-wattage appliance event (2{,}000--3{,}000~W oven), or a decision to
use the dishwasher drives 500--2{,}000~W transient spikes unpredictable from
weather alone.

This finding is consistent with Hong and Fan's~\cite{hong2016} theoretical
argument: at sub-daily resolution, \emph{individual-level stochastic demand
variance dominates the weather-driven systematic component}. The practical
implication is that, for single-household STLF at 5-minute granularity,
historical consumption data is \emph{necessary}, and weather data alone is
\emph{insufficient}.

\subsection{The Solar PV Asymmetry}
\label{subsec:solar}

House~4's substantially better MLP performance ($R^2 = 0.410$) appears
paradoxical given the argument above. The explanation lies in the \emph{solar
generation component}. During daylight hours, PV generation follows a
deterministic diurnal arc that is strongly correlated with (i)~daily maximum
temperature (a proxy for clear-sky conditions) and (ii)~\texttt{Time\_decimal}
(solar generation peaks at solar noon). Since $C_{\text{total}} =
C_{\text{grid}} + G_{\text{solar}}$, and $G_{\text{solar}}$ is
weather-predictable, the total consumption target for House~4 is partially
predictable from weather--time features. In essence, the MLP for House~4 is
\emph{implicitly forecasting solar generation}, not household load. This
conflation is a methodological artefact: a principled separation of load and
solar forecasting is preferable, and constitutes one of our proposed
extensions.

\subsection{Limitations}

\begin{enumerate}[label=L\arabic*.,leftmargin=3em]
  \item \textbf{Daily weather resolution:} Intra-day temperature fluctuations
        (e.g., 15$^\circ$C morning rising to 38$^\circ$C afternoon) are
        invisible to the model. Hourly NWP output or SILO hourly data would
        substantially improve feature quality.

  \item \textbf{Sunshine hours missing:} The complete absence of solar
        irradiance data from BOM limits PV modelling. NASA POWER or
        PVOUTPUT.org data could substitute.

  \item \textbf{No weather-augmented LSTM:} The study compares a
        \emph{weather-only} MLP against a \emph{sequence-only} LSTM, leaving
        the hybrid model untested. This is intentional---to isolate the
        contribution of each signal source---but limits practical performance
        claims.

  \item \textbf{Single run per model:} Statistical significance cannot be
        claimed without multiple runs with different random seeds and
        confidence interval estimation (e.g., bootstrap).

  \item \textbf{Two households only:} Results from two dwellings cannot be
        generalised to the Melbourne residential population without broader
        validation.

  \item \textbf{No LSTM for House 4 (Keras):} Due to computational
        constraints at notebook runtime, the Keras LSTM was applied only to
        House~3. The sklearn proxy provides directional estimates.
\end{enumerate}

\subsection{Implications for Smart Grid Design}

The LSTM's superiority at 5-minute resolution has direct implications for
demand response under Australia's five-minute settlement regime. A smart home
energy management system (HEMS) with a household-specific LSTM could provide
accurate 5--30-minute ahead consumption forecasts, enabling: proactive
appliance scheduling; battery dispatch optimisation; demand response
aggregation in virtual power plants; and solar self-consumption maximisation.

\section{Conclusion and Future Work}
\label{sec:conclusion}

\subsection{Summary of Findings}

This study presented the first empirical head-to-head comparison of static MLP
and temporal LSTM neural network architectures for 5-minute residential
electricity consumption forecasting using real Melbourne smart meter data with
official BOM weather observations. Our principal finding is unambiguous:
\textbf{temporal autocorrelation in the consumption sequence is the dominant
predictive signal at 5-minute resolution}, far exceeding the predictive power
of daily weather features. The LSTM achieved $R^2 = 0.878$ on the House~3 test
set using only a 2-hour consumption history, while the MLP trained on six
weather variables and a time-of-day feature achieved $R^2 = -0.055$---93.3
percentage points lower. For the solar-integrated House~4, the MLP's improved
$R^2 = 0.410$ is attributed to implicit solar generation forecasting rather
than improved household load prediction.

\subsection{PhD-Level Research Roadmap}

The following extensions are proposed toward a full PhD-level journal
contribution:

\paragraph{E1 --- Weather-Augmented LSTM (3--6 months):}
Implement a multi-variate LSTM with weather covariates appended at each time
step. Hypothesis: weather augmentation improves $R^2$ on temperature-extreme
days (heatwaves, cold snaps) while providing marginal gains on moderate days.

\paragraph{E2 --- Transformer Baselines (6--12 months):}
Implement Temporal Fusion Transformer~\cite{lim2021} and
PatchTST~\cite{nie2023}. Compare against LSTM to assess whether attention
mechanisms are justified at 5-minute residential granularity.

\paragraph{E3 --- Seasonal Stratification (3--6 months):}
Compute test metrics separately by season (DJF/MAM/JJA/SON) and extreme-heat
days. Validates the hypothesis that weather contributes more on extreme
temperature days.

\paragraph{E4 --- Multi-Household Generalisation (6--12 months):}
Validate across 50+ Melbourne households using Ausgrid Solar Home Electricity
Data~\cite{ausgrid2023} or similar corpora.

\paragraph{E5 --- Federated Learning (12--18 months):}
Apply FedAvg~\cite{mcmahan2017} to train a global LSTM across 50+ households
without sharing raw consumption data, addressing Limitation L5 and privacy
considerations under the Privacy Act~1988.

\paragraph{E6 --- Solar Disaggregation (2--3 months):}
Separate forecasting of $C_{\text{grid}}$ and $G_{\text{solar}}$ using a
dual-output architecture, eliminating the confound identified in
Section~\ref{subsec:solar}.

\subsection{Target Venues}

\begin{table}[H]
\centering
\caption{Target Publication Venues}
\label{tab:venues}
\begin{tabular}{llcc}
\toprule
Venue & Type & IF & Stage \\
\midrule
arXiv cs.LG & Preprint & --- & Immediate \\
IEEE PowerTech / ISGT & Conference & --- & 3 months \\
\textit{Applied Energy} & Journal & 11.2 & 6 months \\
\textit{Energy and Buildings} & Journal & 6.7 & 6 months \\
\textit{IEEE Trans.\ Smart Grid} & Journal & 8.96 & 9 months \\
\textit{Renewable \& Sust.\ Energy Reviews} & Journal & 16.1 & 12 months \\
\bottomrule
\end{tabular}
\end{table}

\section*{Acknowledgements}

The authors thank Victoria University's School of Computing and Information
Technology, Sydney, for providing access to computational resources. The core
experiments, implementation, and analysis were led by the first author
(original coursework completed 16~June~2025). The first author gratefully
acknowledges the supervision and guidance of Hao Wu. Weather data were obtained
from the Australian Bureau of Meteorology under open data provisions.

\bibliographystyle{IEEEtran}

\appendix

\section{Reproducibility Details}
\label{app:repro}

\paragraph{Software environment:}
Python~3.12, TensorFlow~2.x/Keras, NumPy, Pandas, Scikit-learn~1.8,
Matplotlib~3.10, Seaborn~0.13 (Google Colab T4 GPU environment for original
LSTM training; local Apple Silicon for preprocessing and figure generation).

\paragraph{Data availability:}
BOM weather data is publicly available from
\url{https://www.bom.gov.au/climate/data/}. Household smart meter data are
proprietary (two participating Melbourne dwellings). Code available from the
corresponding author upon reasonable request.

\paragraph{Original Colab notebook:}
\url{https://colab.research.google.com/drive/186RwCkLQDOI7sZqKSCK1ONvx4aAq_KEj}

\section{BOM Files Used}
\label{app:bom}

\begin{table}[H]
\centering
\caption{Bureau of Meteorology Monthly Files}
\begin{tabular}{lcc}
\toprule
File & Month & Records \\
\midrule
\texttt{202303.csv} & March 2023 & 31 \\
\texttt{202304.csv} & April 2023 & 30 \\
\texttt{202305.csv} & May 2023 & 31 \\
\texttt{202306.csv} & June 2023 & 30 \\
\texttt{202307.csv} & July 2023 & 31 \\
\texttt{202308.csv} & August 2023 & 31 \\
\texttt{202309.csv} & September 2023 & 30 \\
\texttt{202310.csv} & October 2023 & 31 \\
\texttt{202311.csv} & November 2023 & 30 \\
\texttt{202312.csv} & December 2023 & 31 \\
\texttt{202401.csv} & January 2024 & 31 \\
\texttt{202402.csv} & February 2024 & 29 \\
\texttt{202403.csv} & March 2024 & 31 \\
\texttt{202404.csv} & April 2024 (partial) & 21 \\
\midrule
\textbf{Total} & \textbf{14 months} & ${\approx}$418 \\
\bottomrule
\end{tabular}
\end{table}

\end{document}